# Intelligent Agent for Hurricane Emergency Identification and Text Information Extraction from Streaming Social Media Big Data

## Jingwei Huang*, Wael Khallouli, Ghaith Rabadi, Mamadou Seck

Department of Engineering Management and Systems Engineering
Old Dominion University
2101 Engineering Systems Building, Norfolk, VA 23529, USA
Email: j2huang@odu.edu
Email: wkhallouli@odu.edu
Email: grabadi@odu.edu
Email: mseck@odu.edu

*Corresponding author

**Abstract:** This paper presents our research on leveraging social media Big Data and AI to support hurricane disaster emergency response. The current practice of hurricane emergency response for rescue highly relies on emergency call centres. The more recent Hurricane Harvey event reveals the limitations of the current systems. We use Hurricane Harvey and the associated Houston flooding as the motivating scenario to conduct research and develop a prototype as a proof-of-concept of using an intelligent agent as a complementary role to support emergency centres in hurricane emergency response. This intelligent agent is used to collect real-time streaming tweets during a natural disaster event, to identify tweets requesting rescue, to extract key information such as address and associated geocode, and to visualize the extracted information in an interactive map in decision supports. Our experiment shows promising outcomes and the potential application of the research in support of hurricane emergency response.

**Biographical notes:**

Jingwei Huang, Ph.D. in Information Engineering (University of Toronto, 2008), is an Associate Professor at Old Dominion University. He has extensive research experience applying artificial intelligence in systems engineering, digital information management, cybersecurity and trust management. His work includes formal semantics based computational trust models, knowledge provenance ontologies, applying computational trust models in public key infrastructures and distributed trust management, access control mechanisms integrating role-based and attribute-based models, and their applications in industry. Dr Huang's current research focuses on Digital Systems Engineering, Big Data, Machine Learning and AI, and Digital Mechanisms of Trust and Security.

Wael Khallouli is currently a Ph.D. candidate and a graduate research/teaching assistant in the department of engineering management and systems engineering at Old Dominion University. He received his B.S. and M.S. degrees in computer science from the University of Tunis. His research interests are broadly centered on big data analytics, digital systems engineering, model-based systems engineering, cloud scheduling, and operations research. His Ph.D. research aims at developing automatic methods to extract actionable information from social media to support disaster response activities. Wael has participated in several funded research projects at Old Dominion University focusing on different topics such as verifiable modeling and simulation in systems engineering and digital twins. Previously, he was a research assistant at Qatar University working on a funded research project focusing on developing analytical tools for logistics planning for Qatar 2022 FIFA world cup tournament.

Ghaith Rabadi is a Professor of Engineering Management & Systems Engineering at ODU. He has Ph.D. in Industrial Engineering with over 20 years of academic and industrial experience. He has published a book, and numerous journal and conference peer reviewed papers. His research has been funded by NATO, NASA, Homeland Security, Virginia Port Authority, Northrop Grumman, MITRE, STIHL, CACI, Sentara Hospitals and Qatar Foundation. His research interests include Planning and Scheduling, Operations Research, Simulation Modeling and Analysis, and Logistics. He is a co-founder and is currently the Chief Editor for the International Journal of Planning and Scheduling. More at ghaithrabadi.com.

Mamadou Seck is an Assistant Professor of Engineering Management and Systems Engineering at Old Dominion University. He graduated in 2007 with a Doctorate in Mathematics and Information Systems at the University of Marseille in France. His research interests are in the fields of system theory, systems engineering, and modelling and simulation, with applications in energy, transportation, and business processes.

---

# 1    Introduction

Hurricane is a seasonal natural disaster, causing the loss of human life, the damages of properties such as buildings and vehicles, and infrastructures destruction such as bridge damage and power outages. Hurricane is listed in the top among the costliest natural disasters in the United States (Statista, 2020). Just list the top three costliest disasters as examples, Hurricane Katrina (2005) cost 161 billion US dollars, Hurricane Harvey (2017) cost $125 billion, and Hurricane Maria (2017) cost $90 billion. Hurricanes count 7 among top 10 costliest natural disasters. It is an important research topic regarding how to ensure society's resilience and infrastructures to hurricanes. This paper presents our research on a proof-of-concept of using an intelligent agent for automatic identification of the emergent requests for disaster rescue from social media data in real-time, to support hurricane emergency response systems. This work is triggered by the observation of the devasting Hurricane Harvey.

Hurricane Harvey made landfall in the southwest coastline of Texas (less than 180 miles to Houston) as a category 4 hurricane with wind speed up to 130 mph (215km/h) on 10 pm August 25, 2017. After the landfall, the hurricane moved very slowly towards northeast inland and dumped a tremendous amount of rainfall (commonly over 40 inches of precipitation, and the maximum up to 60 inches), which caused the catastrophic flooding in Texas, particularly the Houston metropolitan area. In Texas, 336,000 people lost electricity, 300,000 structures and 500,000 vehicles were damaged or destroyed; 185,000 homes were damaged and 9,000 destroyed; tens of thousands people requested rescue, over 13,000 people were rescued; but 107 people died in the storm-related incidents, and the storm is the direct cause of 68 deaths, which is the highest number in Texas since 1919 (Blake & Zelinsky, 2018; National Weather Service, n.d.).

During Hurricane Harvey and the following flooding, many stranded people could not reach 911 and other emergency centres for helps, because those call centres received a surged number of emergency calls, which are much beyond call centres' capabilities to handle. According to a news report (Sedensky, 2017), the 911 call centre at Houston received and processed 75,000 calls in a single day between August 28th to 29th, which is over eight times of normal load per day; and that number does not include the thousands calls that hang up in excessive waiting time. When stranded people could not reach emergency response centres by phone, they turned to social media (such as Twitter, Facebook, Instagram, Reddit, and others) to ask for help.

However, emergency response agencies currently do not have the capacity to monitor and identify emergency messages from a very large volume of streaming social media data. For this reason, phone call is still the only channel to reach emergency response agencies for help. For example, US Coast Guard (USCG) asked people to keep calling emergency centres rather than posting on social media (Shu, 2017; USCG, 2017). To tackle the above situation during Hurricane Harvey, voluntary organizations and many volunteers helped to read through social media and to find posts requesting for helps. Examples include: Digital Humanitarian Network (DHN) sent USCG command centres information every 6 hours in excel spreadsheets with the collected information needed for rescue. Two USCG Academy cadets organized 500 volunteers worldwide to read social media posts, and then they use GIS to create maps for a command centre (Bergman, 2017).

The observation of this hurricane disaster event reveals a critical deficiency of current natural disaster emergency response systems, lacking the capability to monitor and identify emergency messages from the increasingly popular communication channel - social media. At the same time, the event also shows the need and opportunity to develop such capability by leveraging social media Big Data and artificial intelligence.

Our overarching research goal is to develop the technical approach for an intelligent agent that can be used as an automatic tool to enable emergency centres the capabilities of monitoring real-time social media data, identifying emergencies and help requests, making situation-aware decisions in rescue operations, and coordinating and leading crowdsourcing (or volunteering) emergency response efforts.

The contents of this paper are organized as follows. After the introduction of the problem to address and our research objectives in this section, we discuss related research in the next section; we present our technical approach in section 3; we discuss our experiment and evaluation of this work in section 4; finally, we conclude the research and discuss further research in section 5.

## 2    Related Research

Social media platforms, such as Twitter and Weibo, are becoming an important source of real-time information during natural disasters. The research of using social media in the disaster management process is rapidly growing in recent years, particularly with the fast spread of deep learning technologies.

Early research studies proposed statistical models for disaster events detection and location estimation. For instance, Sakaki et al (Sakaki, Okazaki, & Matsuo, 2010) proposed an earthquake warning system using spatial and temporal information from Twitter to detect critical events (e.g., earthquakes and typhoons) occurrences in real-time. They trained an SVM classifier to detect tweets related to a target event. The classifier employs three feature categories: special keywords (e.g., earthquake), statistical features (e.g., number of words), and the context surrounding the keywords. The authors proposed a spatio-temporal statistical model for detecting the target events and locating their centers and trajectories. Kalman and Particle filtering algorithms were applied for location estimation. An exponential distribution was used to fit data (users posts' times) and detect the disasters. Earle et al. (Earle, Bowden, & Guy, 2011) proposed a time-series procedure for early earthquake detection from Twitter messages. Cheng et al. (Cheng & Wicks, 2014) proposed a space-time scan statistics (STSS) technique to cluster tweets across space and time dimensions for the task of disaster events detection.

Yuan and Liu (Yuan & Liu, 2018) applied data mining and semantic analysis methods to identify affected areas during major disasters from social media platforms (Twitter) and support damage assessment activities. In this paper, the authors employed a keyword-based approach for filtering disaster and damage-related tweets. Disaster tweets are filtered using a manually created index dictionary which comprises the most common keywords and expressions for describing damages. Caragea et al. (Caragea, Silvescu, & Tapia, 2016) proposed a convolutional neural network model (CNN) to classify tweets into informative (i.e., crisis-related) and non-informative tweets. With the proposed CNN, they evaluated on 6 datasets from different disaster events and outperformed the baseline models, such Naïve Bayes (NB), support vector machine (SVM), and a 3-layer artificial neural network (ANN). Nguyen et al. (Nguyen et al., 2017) also proposed a CNN model to classify tweets into informative and non-informative tweets. Different from the previous model (Caragea et al., 2016) that uses a simple bag-of-word encoding to represent input data, this model uses two word embedding models: word2vec (Mikolov, Chen, Corrado, & Dean, 2013) and a domain-specific word embedding trained on disaster-related corpus. The model was evaluated in two different scenarios: (i) in-domain classification where it was trained and tested on the same disaster event data, and (ii) out-domain classification where the model was trained on one disaster event but tested on unseen data from another event. The proposed CNN provided better performance in both cases compared to baseline classifiers, such as support vector machine (SVM), logistic regression (LR), and random forest (RF). Sit et al. (Sit, Koylu, & Demir, 2020) proposed an integrated framework that detects informative tweets (binary classification), and then applies an unsupervised topic modelling approach to extract relevant topics (e.g, affected individual, donation, and caution and advice) from the content. The authors evaluated 2 deep learning models: a convolutional neural network (CNN), and a bidirectional long-short term memory (LSTM). The authors found that both LSTM and CNN models outperform the baseline models (SVM and LR) with the LSTM model slightly outperforming the CNN in accuracy and F1 score metrics, in the experiments with the data collected during Hurricane Irma. Kabir and Madria (Kabir & Madria, 2019) proposed a system that automatically classifies Tweets into multiple categories – i.e., rescue needed, water needed, injured, sick, flood, and disabled elderly children and women. A deep learning approach that combines convolutional layers, LSTM layers, an attention layer, and an auxiliary feature layer that integrates semantics. The model was evaluated on both in-domain and out-domain cases using CrisisNLP (Imran, Mitra, & Castillo, 2016) and CrisisLex (Olteanu et al, 2014) data. Those research shows that deep learning is a promising approach to go. However, to use deep learning models, which are typically supervised machine learning, it is necessary to have the labelled dataset for training. It is a labor-intensive and time-consuming process to do data labelling. Additionally, deep learning may suffer from low generalization between different event types. More recent research employed transformer-based classification models, such as Google BERT, for the disaster classification task. For instance, Liu et al. (Liu, Blessing, Lucienne, Wood, & Lim, 2020) proposed an end-to-end BERT-based classifier, called CrisisBERT, for two relevant disaster management tasks: (1) crisis detection (i.e., identifying crisis-related tweets), and crisis recognition (i.e., categorizing tweets into multiple informative classes). The proposed model outperforms the state-of-the-art models using multiple data benchmarks. Zahera et al. (Zahera, Elgendy, Jalota, & Sherif, 2019) proposed a fine-tuned BERT model to classify disaster tweets into 25 information categories, such as search and rescue, donations, weather, and emerging threats. The authors evaluated two BERT variants using different loss functions (binary cross-entropy and focal loss). Fan et al. (Fan, Wu, & Mostafavi, 2020) proposed a hybrid machine learning pipeline for detecting relevant disaster tweets and their locations from the content. The authors employed a BERT-based classifier to categorize tweets into multiple humanitarian categories. In this study, the BERT classifier showed a better performance than the baseline models including GRU, LSTM, and CNN deep learning models. Madichetty (Madichetty, Muthukumarasamy, & Jayadev, 2021) proposed a multi-modal approach to identify informative tweets including both images and text. The authors used a combination of BERT-based model for text classification and densely connected convolutional network (DenseNet) model for image classification.

Identification of location information is vitally important for disaster emergency response. Although Twitter provides many location fields, such as users' locations, place names, and geo-coordinate, there are many limitations for various reasons. Critically, social media users generally tend to not attach their geo-location information due to privacy concerns. According to research, only 1 to 3% of tweets are geotagged (Zahra, Imran, & Ostermann, 2018). Researchers proposed many techniques including deep learning to estimate users' locations either from the content they post or from their historical data.

Lau et al. (Lau, Chi, Tran, & Cohn, 2017) proposed an end-to-end deep neural network, called 'deepgeo', that predicts the location of a tweet from a total of 3362 cities provided in the training set. The proposed deep learning model uses several features from the tweet meta-data, including textual content, creation time and user account creation time. Snyder et al. (Snyder, Karimzadeh, Chen, & Ebert, 2019) improved 'deepgeo' model by adding word embeddings (e.g., word2vec). The new model outperformed the original one by 3% accuracy. Kumar and Singh (Kumar & Singh, 2019) trained a convolutional neural network to detect location words in disaster-related tweets. They collected data during multiple earthquake events from different countries. The problem was formulated as a multi-label classification problem, where the model's output is a zero-one vector – i.e., location words are encoded as 1 while non-location words as 0. The proposed model achieved significant accuracy in recognizing location words at several granularity levels (e.g., country names, city names, and street names). After testing several CNN configurations, the best configuration achieved an F-score equal to 95.3%. The above works provide helpful clues, but they are in a large granularity of location identification such as city-level. In our research, we need much finer granularity of location identification at house level.

Several studies used Named Entity Recognition (NER) to detect location entities in the tweets. For instance, Fan et al. (Fan et al., 2020) implemented a Named Entity Recognition (NER) tool to detect location references mentioned in the emergency tweets. The identified locations are matched to existing locations and place names in the disaster area using a Google map API. The limitation of this approach is the need to create location references.

Singh et al. (Singh, Dwivedi, Rana, Kumar, & Kapoor, 2019) used Markov-Chain based inference method to estimate the locations of users posting help requests during natural disaster based on their historical posts. Their system first detects high-priority tweets – i.e., tweets from users' asking for help – and then forward geo-tagged tweets to the rescue teams. Tweets lacking geotagging information are further processed. Their locations are inferred using the users' location in the last 7 days. Historical locations are modelled as a sequence of events using Markov Chain model. The proposed location detection algorithm achieved 87% accuracy, as reported by the authors. It is an interesting approach to use historical data to infer location.

The above research works provide important information in the process of our research. However, most of them are in supervised machine learning approach. Before we can leverage machine learning technologies, first, we need to have an approach to start with unlabeled raw data collected during Hurricane Harvey.

## 2.1 Contributions

The contributions of this paper are twofold. First, we identified a major deficiency of the current Hurricane disaster emergency response system in the US, that is, the limited and non-scalable capacity of emergency hotline service and lack of capability to monitoring social media. For this deficiency, we proposed a solution and developed a proof-of-concept of using an intelligent agent as an automatic tool with the capabilities of monitoring real-time social media data, identifying emergency rescue requests, and extracting information to support emergency response. Secondly, we developed a logic-based approach to dealing with unlabelled Twitter data, identifying rescue request tweets, and extracting relevant information from those tweets. The research progress of deep learning and many well-developed Machine Learning APIs and toolsets make machine learning broadly applied in many fields, including text classification like identifying a targeted class of tweets that we address. However, classification using supervised learning has three general issues: (1) data must be labelled; (2) the trained model could be sensitive to the training dataset, or the predictions with the model are dependent on the training dataset; (3) dependent on the type of the ML models to use, typically, neural network models suffer from the explainability problem. We address these issues by using a logic model based on the logical structure of the problem. By this approach, the outcome is explainable, modelling does not rely on labelled data, and the model is independent of the dataset used for verification.

## 3 Technical Approach and Experiments

In this research, we select Twitter as a representative source of social media big data, and we develop our technical approach and the proof-of-concept solution and conduct experiments with R. R is one of most powerful languages and development environments for Data Science. Twitter is one of the most popular social media, and Twitter offers APIs to enable developers to collect both streaming twitter data at real time and historical tweets for a period of time in the past.

To achieve the research goal presented in section 1, we conducted the following main aspects of research: collecting streaming tweets at real time, building logical filters to capture the features of tweets, using logical filters to identify tweets that request hurricane emergency rescue, locating a rescue place with geocode, translating non-English tweets into English for information extraction, and plotting and posting information of each rescue request in an interactive map. The overview of our technical approach and workflow is illustrated as figure 1. We discuss each major aspect of research as the following subsections.

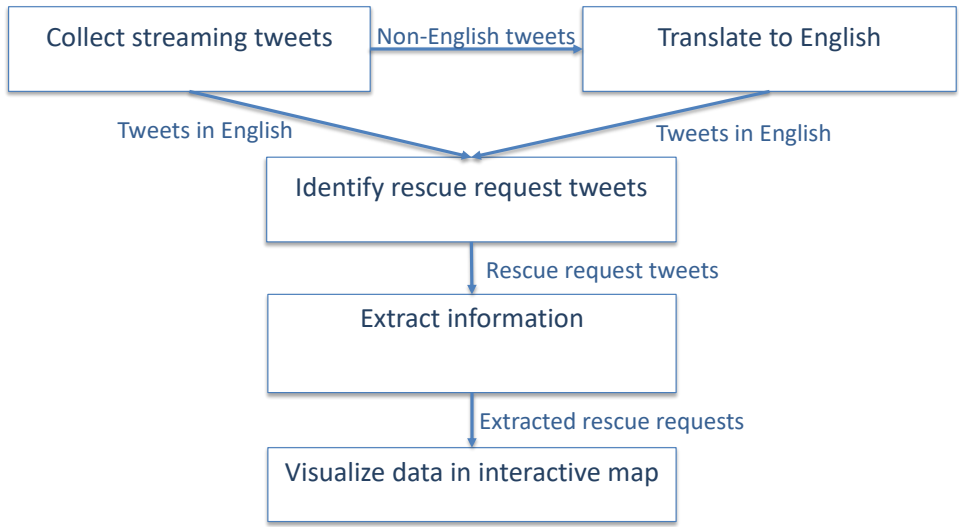

Figure 1. Workflow of the prototype of the intelligent agent for hurricane emergency identification and text information extraction from streaming social media Big Data.

### 3.1 Collecting streaming tweets at real time

Twitter actively offers very useful public APIs to access both recent historical tweets and the streaming tweets at the real time. We collected the streaming tweets from August 26 to August 31, 2017 during Hurricane Harvey. Twitter API allows to collect streaming tweets with several operators including keywords, Hashtags, bounding _box, and many others. The following keywords and hashtags, "#HurricaneHarvey", "#Harvey", "Hurricane",

"flooding", were used with logical relationship OR, and the bounding box we used is (-99, 27.6, -90.8, 33.5) as illustrated in figure 2. The logical relationship between keywords and bounding_box is fixed as OR in Twitter API. In R, both packages "streamR" and "rtweet" facilitate collecting the streaming tweets. In our research, "streamR" was used and it has demonstrated its very stable performance that it is able to automatically reconnect and collect tweets if the connection is dropped. We use "rtweet" to read in stream tweets. The package facilitates to read the expanded 280 characters.

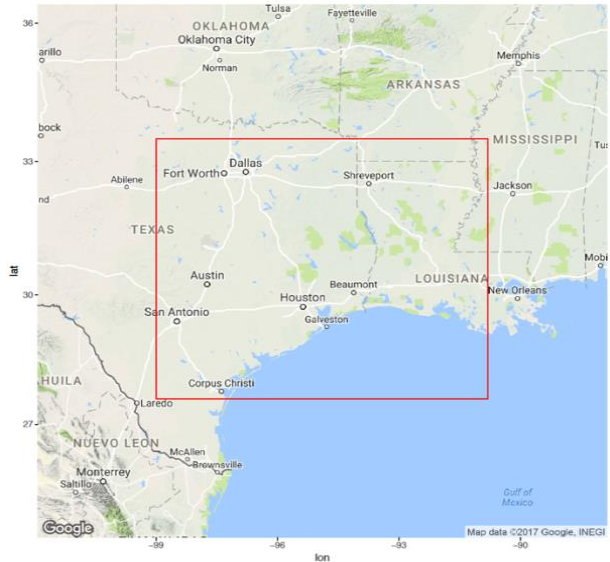

Figure 2. The bounding box used to collect streaming tweets.

### 3.2 Logic-based approach for identifying tweets requesting rescue

Because the tweets we collected during Hurricane Harvey are raw data and there is no labelled data for machine learning, we adopt a logical approach to identifying emergency rescue requests from streaming tweets, by using *regular expressions* (Regex) to catch the language patterns or features of those tweets. Let us start with the distinct features of tweets requesting rescue as follows.

**Feature of containing address of street number**

First of all, it is necessary for any request for rescue to contain their location information. During hurricanes, emergency response agencies constantly urged people to provide their precise address when requesting helps, most people indeed did so, and only a small number of people described their location with street features such as "in the intersection of x and y", "in the end of road x", and so on. Any requests of rescue without specific location information will only make it an ineffective request and not helpful. Location information is a necessary component of all rescue requests.

Addresses in the US have a very clear and simple pattern as follows:

<House Number> <Street Name> <Street Suffix> [<Unit>]
<City/Town Name> <State> <Zip code>

*Intelligent Crowdsourcing Hurricane Emergency Response Support System*

This pattern can be captured with Regex. To maximally capture all tweets mentioning address, we start with selecting all tweets matching linguistic part of <House Number> <Street Name> <Street Suffix> from text. <House Number> is typically an integer with limited length. In our search, the highest house number in the US we found so far is 107900 in address "107900 Overseas Hwy, Key Largo, FL 33037". In our regex, we match house number as one to six digits. <Street Name> is a number of words, which can have large variety even though some patterns exist. The very feature of a <Street Name> is that it is followed with a <Street Suffix>, such as "Street", "Ave" and others. We use the street suffixes listed in (Wikipedia, 2020). There are more than two hundred street suffixes used in the US. The regex for address is created as follows.

```
rgx.address <- "\\b\\d{1,6}\\s+(#?[A-z]+\\.?(-[A-z]+)?\\s+){1,3}\\b(#@#
|Alley|Allee|Ally|Aly#@#
|Annex|Anex|Annx|Anx|#@#
|Arcade|Arc|#@#
|Avenue|Av|Ave|Aven|Avenu|Avn|Avnue|#@#
...
|Well|Wl|Wells|Wls|#@#
).?\\b"
```

This regex is very long, and 125 lines in the middle are omitted. In the regex, substring "\\b" is an anchor matching word boundary; substring "\\d{1,6}" matches 1 to 6 digits; "\\s+" match one or more spaces including character for space, tab, vertical tab, newline, carriage return, and form feed. Substring "#?" is for the optional hashtag which is possibly used in street names by some twitter users, for example, "3 friends stuck at 4055 South #Braeswood Boulevard …". For general application (rather than specifically for Tweets), substring "#?" should be removed. Together, substring "(#?[A-z]+\.?(-[A-z]+)?\\s+){1,3}" represents one to three words consisting of English characters, possibly connected with "-" or ended with ".". Substring "#@#" followed by a new line matches exactly "#@#\n", which is assumed not present in the text to be scanned, and the purpose of this substring is to introduce a way to write a very long regex in multiple lines. "Alley", "Allee", "Aly", and others are possible street suffixes. Because there are more than two hundred street suffixes, in the piece of Regex shown above, we only show the first few lines and last line of street suffixes, most lines in the middle are omitted.

Addresses may appear in another form such as "1108 Highway 7", "123 Ave. G", and others. This form of addresses is matched with the following regex.

```
rgx.address2 <-"\\b\\d{1,6}\\s+(AVENUE|AV|AVE|AVEN|AVENU|AVN|AVNUE|#@#
|HIGHWAY|HWY|HIWAY|HIWY|HWAY|HWY|#@#
|ROAD|RD|ROADS|RDS|#@#
|ROUTE|RTE|#@#
|STREET|ST|STRT|STR|STREETS|STS|#@#
).?(\\d+|[A-z])\\b"
```

In our current stage of research, we left location description such as "the intersection of street x and y" for future work.

**Feature of containing words for asking helps**

Location information is a necessary feature for hurricane rescue request tweets; however, it is not sufficient. To capture rescue requests, some other features need to be identified. Another relevant feature is whether a tweet contains hashtags or key words about requesting for helps such as "HurricaneRescue", "FloodRescue", "please help", "need to be rescued", and so on. This feature can be captured easily with a regex matching corresponding key words.

**Feature of containing contextual information**

Some rescue requests contain context information. We identify three types of text information, including:

- The hashtags or names used to address the disaster, such as "Hurricane Harvey", "HurricaneHarvey", "HurricaneFlood", "HoustonFlood", and so on.
- The words about a disaster and a region, such as {"Texas", "Harvey"}, {"Bay City", "Flood"}, {"Houston", "Flood"}, and others.
- The words about situation description, such as "stranded", "stuck", "trapped", "rooftop", "attic", and others.

**Features of tweets that are not rescue requests**

Some tweets have the above features related to hurricane disaster, but they are not rescue requests, instead they are about to update rescue status, to offer helps for people who need shelter, foods, and others, to express hurricane response related pollical opinions, to report hurricane related news by media or news agencies, and to make commercial advertisements. Those non-rescue requesting tweets have certain patterns or features can be captured with regex either.

**Integrating filters for identifying tweets requesting rescue**

The features addressed above can be captured by our regex filters. Finally, we can integrate those filters to capture the tweets requesting rescue. The logical relationship between those filters can be expressed in the First-Order Logic as follows.

$$
\begin{aligned}
\forall x \, ( & \text{hasFeatureAddress}(x) \land \\
& (\text{hasFeatureAskHelp}(x) \lor \text{hasFeatureDisasterContext}(x)) \land \\
& \neg \, (\text{hasFeatureStatusUpdate}(x) \lor \text{hasFeatureOfferHelp}(x) \lor \\
& \qquad \text{hasFeatureNewsReport}(x) \lor \text{hasFeaturePolitical}(x) \lor \\
& \qquad \text{hasFeatureAds}(x)) \\
& )
\end{aligned}
$$

where the universe of discourse is the tweets collected; the semantics of each predicate is as indicated as their names and those features which we have discussed in this section.

### 3.3 Information Extraction

After rescue request tweets have been identified, rescue request information particularly specific location including full address and geocode need to be extracted, to support search and save operations of disaster emergency response. The challenge is that many tweets did not provide standard formats of addresses and frequently missed some components of an address. In order to catch as more address components as possible, we maximize the flexibility of full address pattern design. We decompose full address pattern as following components:

*<House Number>*
*( (<Street Name> <Street Suffix>) | (<Street Suffix> <Street Character or Number>) )*
*[<Unit>]*
*[ [#]<City/Town Name> ]*
*[<State> ] [<Zip code>]*

Where a square bracket [<...>] is used to denote an optional pattern, "#" is possibly before city name, used as a hashtag, and between address components, a "connector pattern" is inserted. A connector could be one or more characters for a "space", a "comma", a "period", a "tabulator", a "newline" or a "carriage return".

If the address information of a rescue request tweet is incomplete, we assume that the request is from Houston or Texas, which is the central of the disaster event. So, in the case where all optional components are absent in a rescue request tweet, we search whether there exists a hashtag containing a substring of "houston", e.g. "#houstonflooding"; if it is true, we concatenate a string ", Houston, TX" in the end of the extracted full address, otherwise, concatenate a string ", Texas". In the case that the full address patten catch information on city or state, we search the extracted full address to see if it contains Texas; if not, we concatenate a string ", Texas" in the end of the extracted full address.

After the "full" address of a rescue request tweet is extracted, we use it as the search string to call Google Map API and acquire the geocode of the address. Then we plot and post the information of this rescue request tweet in an interactive map, which will be discussed in the next subsection.

### 3.4 Data Visualization for Emergency Response

To support hurricane emergence response to the requests for rescue, we use interactive map to display the inform extracted from the identified tweets of rescue request. Thanks to package "leaflet", which makes the visualization very simply, efficient and effective. On the leaflet interactive map, clicking a marker, which corresponds to the contents extracted from a rescue request tweet, the corresponding contents will be displayed in a pop-up window, as illustrated in the screen shot shown in figure 3. The interactive map can be zoom in and out. The requesting time is saved in the "created_at" attribute of a tweet. The original tweets provide creating time in UTC. The time shown in the screen shot is US/Central time (CDT), which has been transformed from original tweet to reflect the local time when the tweet was created.

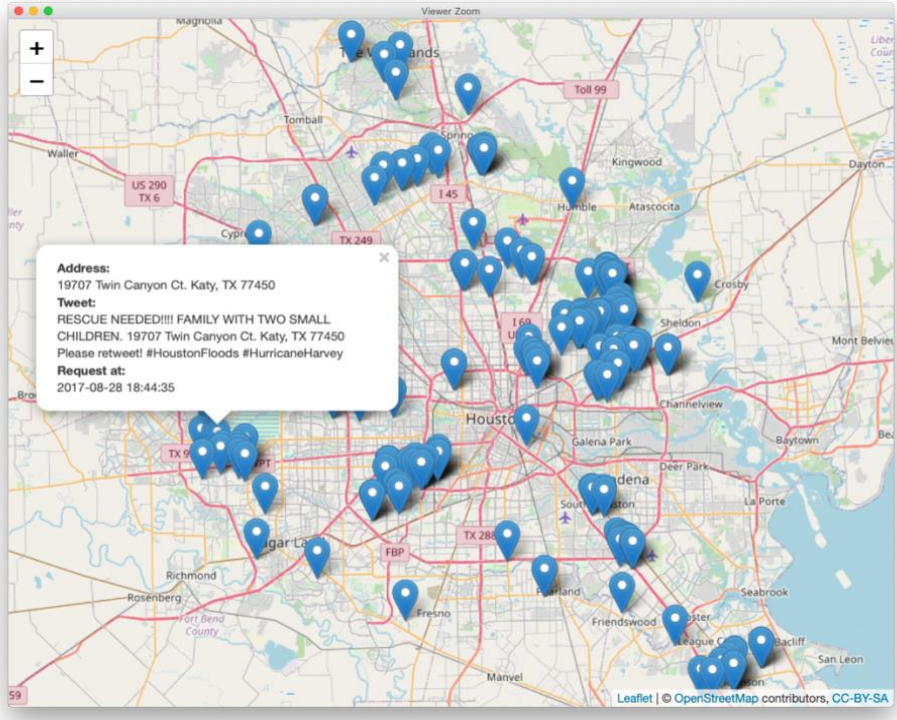

Figure 3. A screen shot of interactive map fed with rescue request tweets

## 4   Experiment and Evaluation

Our prototype system is implemented with R and RStudio, as well as many open-source packages, such as "streamR", "rtweet", "tm", and "leaflet". The system can be deployed on both a laptop or cloud-based on-demand services, so that the system can run on different platforms and leverage the advantages of both the low-cost and flexibility of laptops and the high-performance computing of cloud platforms for Big Data, with respect to computing power and storage scalability. In our experiment, we use both platforms for collecting streaming tweets, data processing, and information extraction. In the scenario of large volume of tweets, Apache Flume is used to directly store the data into our Hadoop cluster, and big data can be processed on Hadoop.

To evaluate the effectiveness of the prototype system, we applied the logic-based filter, which was presented in subsection 3.2, to a selected small set of 5792 tweets (including 251 rescue request tweets), which has been manually labelled incrementally by the graduate students in their coursework on logical filter development and exercise, also has been verified and updated by us. The confusion matrix of the classification is shown in Table 1. From the table, the true positive rate (sensitivity) can be calculated as 0.908, true negative rate (specificity) as 0.988, Matthews Correlation Coefficient (MCC) as 0.832, and F1 score as 0.834. Tweets with false negative error are typically those who did not provide sufficient information of address, such as "@KPRC2 there are stranded families at Creech Elementary on Mason Rd. You have boats nearby. Please send them!". We will address this issue in future work. The outcome of this experiment shows its potential usefulness in support hurricane emergency response.

Table 1. Confusion matrix of experiment result

|  |  | Reference | |
| --- | --- | --- | --- |
|  |  | Positive | Negative |
| Predication | Positive | 228 | 66 |
|  | Negative | 23 | 5475 |

## 5 Conclusion and Further Research

In this paper, we presented our research work on leveraging social media Big Data and AI in support of hurricane disaster emergency response to overcome the deficiencies of the current practice, caused by limited capacity and scalability of emergency call centres and lack of capability to monitor social media. Hurricane Harvey and the following Houston flooding were used as the motivating scenario in our research and the development of a prototype system for the intelligent agent. With the system, we collected streaming tweets in real-time during Hurricane Harvey, developed a logic-based approach to identifying rescue request tweets and extract key information such as address and geocode from each identified tweet, and visualized the extracted information of each rescue request in an interactive map to support decision making. Our experiment shows promising outcomes and the potential application of the research as an intelligent tool for emergency centres in support of hurricane emergency response. The technical approach used in our research is beyond hurricane tweets and can be used for other text information identification and extraction. The proposed logic-based approach to identifying targeted categories of tweets can be applied to unlabelled data, which is necessary for classification and other supervised machine learning approaches.

In the context of Hurricane emergency response in the US, the disaster areas frequently have people who speak Spanish. For a specific engineering solution, we recommend including Spanish keywords corresponding to the English ones to avoid potential inaccurate translation, thus maximizing the chance for saving lives.

Future research can go in several directions. We will leverage the small core set of labelled data and apply a machine learning approach, to continue growing the labelled dataset, to explore an integrated approach and more efficient and effective methods for text information classification and information extraction.

Generally, Twitter is always a morass of information mixed with uncertain, incomplete, inaccurate, or even false or fake information. The trustworthiness of information is an essential topic for further research. From the perspective of security and societal resilience, it is essential to have the capability to identify fake information or misinformation. The research in this direction will need the ground truth which validity is supported with certain evidence. The information trustworthiness can be judged based on semantically linked other posts or reports, the provenance (Baeth & Aktas, 2019; Fox & Huang, 2004), the trust in information sources, directly or derived from a trust network (Huang & Nicol, 2009). In the context of hurricane emergency response research, from our observations of the twitter data, information uncertainty comes from two primary sources. The first one is the fuzziness of the "category" of rescue request tweets. Some news reports from media and disaster situation descriptions from the crowd possibly inform the situations where people need help. We recommend marking out and reporting these types of information for human experts to investigate. The second one is incomplete and/or inaccurate information. Some are due to the language features of tweets, and they can be addressed by text normalization. Basically, incomplete information is a challenging issue. Theoretically, information fusion from other relevant tweets can lead to more information. For more sophisticated AI solutions, instead of treating each case individually, a "possible world" model could be constructed based on the information from social sensors - hurricane-related tweets, to connect and update reported events, and to infer beliefs.

## Acknowledgement

The research work presented in this paper is partially supported by the National Science Foundation under Grant No. 1828593. As an essential component of a larger effort that won the first NATO Innovation Challenge Award, we also express our thanks to NATO Innovation Hub. The authors sincerely thank the students of ENMA 854 Big Data Fundamentals and ENMA 646 Information Systems in 2020 fall semester for their contributions in labelling data. This paper is a revised and expanded version of an invited talk on Social Media Big Data Support for Emergency Response and Logistics at NATO TideSprint 2017, Virginia Beach, USA, 23-27 Oct 2017.